\documentclass[runningheads]{llncs}
\usepackage{graphicx}
\usepackage{amsmath}
\usepackage{amssymb}
\usepackage{mathtools}
\usepackage{cellspace}
\usepackage{multirow}
\begin{document}
\title{Interpreting Context of Images using Scene Graphs}
\author{Himangi Mittal\inst{1} \and
Ajith Abraham\inst{2} \and
Anuja Arora\inst{1}}
\authorrunning{Himangi et al.}
\institute{Department of Computer Science Engineering, Jaypee Institute of Information Technology, India \and
Machine Intelligence Research Labs (MIR Labs), Auburn, WA 98071 USA, Washington United
% States
\\
\email{himangimittal@gmail.com}, \email{ajith.abraham@ieee.org},
\email{anuja.arora@gmail.com}}
\maketitle              % typeset the header of the contribution
\begin{abstract}
Understanding a visual scene incorporates objects,
relationships, and context. Traditional methods working on
an image mostly focus on object detection and fail to
capture the relationship between the objects. Relationships
can give rich semantic information about the objects in a
scene. The context can be conducive in comprehending an
image since it will help us to perceive the relation between
the objects and thus, give us a deeper insight into the image.
Through this idea, our project delivers a model which
focuses on finding the context present in an image by
representing the image as a graph, where the nodes will the
objects and edges will be the relation between them. The
context is found using the visual and semantic cues which
are further concatenated and given to the Support Vector
Machines (SVM) to detect the relation between two objects.
This presents us with the context of the image which can be
further used in applications such as similar image retrieval,
image captioning, or story generation.

\end{abstract}
\section{Introduction}
% \subsection{A Subsection Sample}
Computer Vision has a number of applications which
needs special attention of researchers such as semantic
segmentation, object detection, classification, localization,
and instance segmentation. The work attempted in the paper
lies in the category of semantic segmentation. Semantic
segmentation has two phases – segmentation, detection of
an object and semantic, is the prediction of context.

Understanding a visual scene is one of the primal goals
of computer vision. Visual scene understanding includes
numerous vision tasks at several semantic levels, including
detecting and recognizing objects. In recent years, great
progress has been made to build intelligent visual
recognition systems. Object detection focuses on detecting
all objects. Scene graph generation ~\cite{bib1}~\cite{bib2}~\cite{bib3}~\cite{bib4} recognizes
not only the objects but also their relationships. Such
relationships can be represented by directed edges, which
connect two objects as a combination of the subject –
predicate - object. In contrast to the object detection
methods, which just result in whether an object exists or not,
a scene graph also helps in infusing context in the image. For example, there is a difference between a man feeding a
horse and a man standing by a horse.

This rich semantic information has been largely unused
by the recent models. In short, a scene graph is a visually
grounded graph over the object instances in an image, where
the edges depict their pairwise relationships. Once a scene
graph is generated, it can be used for many applications.
One such is to find an image based on the context by giving
a query. Numerous methods for querying a model database
are based on properties such as shape and keywords have
been proposed, the majority of which are focused on
searching for isolated objects. When a scene modeler
searches for a new object, an implicit part of that search is a
need to find objects that fit well within their scene. Using a
scene graph to retrieve the images by finding context has a
better performance than comparing the images on a pixel
level. An extension to the above application is clustering of
similar images. Recent methods cluster the image by
calculating the pixel-to-pixel difference. This method does
not generalize well and works on images which are highly
similar. Also, this method may lead to speed and memory
issues. The approach of scene graph infused with context
can help to cluster the images even if there is a vast pixel
difference. This method is also translation invariant,
meaning, that a girl eating in the image can be anywhere in
the image, but the context remains the same. Since this
method uses semantic information, it enhances speed and
memory.

The paper is structured in the following manner - Section II
discusses the related work done in this direction,
highlighting the scope of work to design a better solution. Importance and significance of work are
discussed in section III. Section IV is about the dataset
available and used to perform experiments. Section V
discusses the solution approach followed by section VI
which covers finding of object and context interpretation
using scene graph. Finally, concluding remark and future
scope is discussed in section VII.

\section{Related Work}
The complete work is can be divided into two tasks – Object
detection and Context interpretation. Hence, a plethora of
papers have been studied to understand the various
approaches defined by researchers in order to achieve an
efficient and scalable outcome in both directions. Initially,
in order to get an idea about deep learning models used in
the field of computer vision, paper ~\cite{bib6} is studied. This paper
~\cite{bib6} covers the various deep learning models in the field of
computer vision from about 210 research papers. It gives an
overview of the deep learning models by dividing them into
four categories - Convolutional Neural Networks, Restricted
Boltzmann Machines, Autoencoder, and Sparse Coding.
Additionally, their successes on a variety of computer vision
tasks and challenges faced have also been discussed.

In 2016, Redmon et. al. ~\cite{bib5} delivers a new approach, You
Look Only Once (YOLO) for object detection by expressing
it as a regression problem rather than a classification
problem. It utilizes a single neural network which gives the
bounding boxes coordinates and the confidence scores.
Detection of context in images is an emerging application.
Various methods ranging from scene graph to rich feature
representation have been employed for the same. In 2018,
Yang et. al. have developed a model Graph RCNN ~\cite{bib4}
which understands the context of the image by translating
the image as a graph. A pipeline of object detection, pair
pruning, GCN for context and SGGen+ has been employed.
Similar sort of work is done by Fisher et. al. ~\cite{bib7}, they
represent scenes that encode models and their semantic
relationships. Then, they define a kernel between these
relationship graphs to compare the common substructures of
two graphs and capture the similarity between the scenes.

For effective semantic information extraction, Skipgram ~\cite{bib8}
model has been studies works for learning high quality
distributed vector representation. In addition to this, several
extensions ~\cite{bib9} of Skipgram have been experimented with to
improve the quality of vectors and training speed. Two
models have been proposed in the work ~\cite{bib10} which is an
extension to Word2vec to improve the speed and time. Their
architecture computes continuous vector representations of
words from very large data sets. Large improvements have
been observed in the accuracy at a much lower
computational cost. The vectors are trained on a large dataset
of Google for 1.6 billion words. In 2018, a new method
“deep structural ranking” was introduced which described
the interactions between objects to predict the relationship
between a subject and an object. Liang et.al ~\cite{bib11} makes use
of rich representations of an image – visual, spatial, and
semantic representation. All of these representations are
fused together and given to a model of structural ranking loss
which predicts the positive and negative relationship
between subject and object.

The work ~\cite{bib12} aims to capture the interaction between
different objects using a context-dependent diffusion
network (CCDN). For the input to the model, two types of
graphs are used - visual scene graph and semantic graph.
The visual scene graph takes into account the visual
information of object pair connections and the semantic
graphs contain the rich information about the relationship
between two objects. Once the features from visual and
semantic graphs are taken, they are given as an input to a
method called Ranking loss, which is a linear function.
Yatskar et. al. work ~\cite{bib13} is an extension to the predicate and
relationship detection. It introduces a method where it
focuses on the detection of a participant, the role of the
participants and the activity of the participants. The model
has coined the term ``FrameNet'' which works on a dataset
containing 125,000 images, 11,000 objects, and 500
activities.

\section{Importance and Significance of Work}
This work is having its own importance and significance in
varying application due to the following:

\begin{itemize}
    \item An extension to the object detection by finding the
underlying relationship between object and subject.
Object detection merely works on the presence of the
objects giving us partial information about the images.
Context can give us the true meaning of the image.
    \item Classifies the image as similar on the basis of the
underlying context. Object detection classifies the
images as similar on the basis of the presence of
specific objects. However, the images can be quite
different than each other based on context.
Incorporating the context will give a deeper insight into
an image.
    \item If the context is employed on prepositions as well as
verbs (future work), rich semantic information can be
used to generate interesting captions and stories related
to images.
    \item No pixel-to-pixel level similarity/clustering calculation.
One of the applications of incorporating context is to
find similar images. Conventional techniques involve
pixel by pixel calculations, thus increasing the
overhead. Scene graphs save time by considering the
visual and semantic features.
    \item Useful in query processing, image retrieval, story
generation, image captioning. Once the context is
detected, it can be used in various applications like
query processing in search engines, image retrieval
using captioning [14][15], as well as story generation.
\end{itemize}

\section{Datasets}
Most famous datasets used for Scene understanding
applications are MS-COCO ~\cite{bib16}, PASCAL VOC , and
Visual Genome, and Visual Relationship Detection- VRD.

VRD dataset contains 5000 images, 100 object
categories, and 70 predicates. It is most widely used for the
relationship detection for an object pair in testing since it
contains a decent amount of images. COCO ~\cite{bib16} is large scale object detection, segmentation, and captioning dataset.
This dataset is used in several applications- Object
segmentation, Recognition in context, Super pixel stuff
segmentation. It has over 330K images (200K labeled), and
80 object categories. Also, it has 5 captions per image which
can be used for image captioning methods.

To perform the experiments, VRD dataset has been
taken. Visual Relationship Detection (VRD) with Language
Priors is a dataset developed by Stanford aiming to find the
visual relationship and context in an image. The dataset
contains 5000 images with 37,993 thousand relationships,
100 object categories and 70 predicate categories connecting
those objects together. Originally, in the dataset, we are
given a dictionary file of training and testing data which we
convert into training set with 3030 images, test set of 955 images, and validation set of 750 images.

Statistics of the number of objects and visual relationships
in every image is shown in Fig.~\ref{fig1} and Fig.~\ref{fig2}
respectively. In Fig.~\ref{fig2}, the file with an unusual number of
134 relationships in image '3683085307.jpg'.

\begin{center}
\includegraphics[width=8cm, height=6cm]{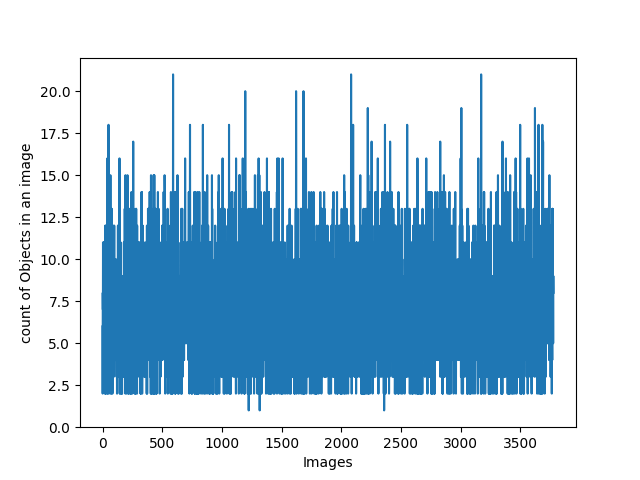} \\
\textbf{Fig.1.} \text{Statistics of the number of objects in images}
\label{fig1}
\end{center}

\begin{center}
\includegraphics[width=8cm, height=6cm]{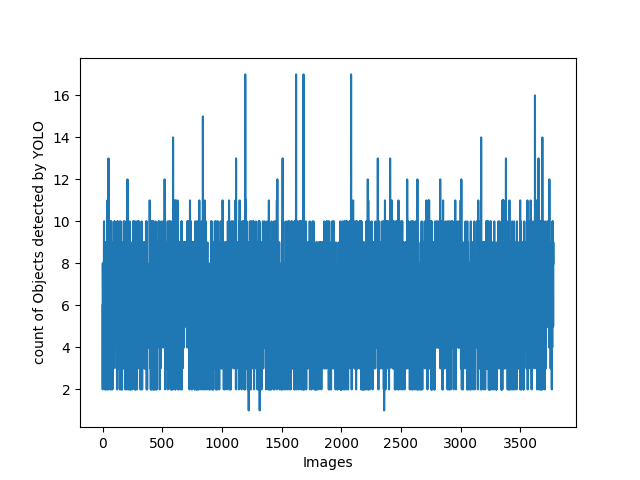}\\
\textbf{Fig.2.} \text{Statistics of the number of objects detected by YOLO}
\label{fig2}
\end{center}

\section{Solution Approach}
The purpose of this project is to extend the object
detection techniques to find context and hence, understand
the underlying meaning of an image. Our idea uses 2D
images of various scenes which have objects interacting
with each other. The interaction can be in the form of
prepositions like (above, on, beside, etc) or activity form
like (cooking, eating, talking, etc). The model considers the
scenery. The solution approach is basically divided into four
modules. These modules are clearly depicted in Fig.~\ref{fig3}.
The first phase is the object detection for which YOLO
object detector has been used. YOLO will provide an image
with a bounding box of detected objects. This will be used
to identify the semantic and visual features of the image.
VGG-16 is used to generate visual features and Word2Vec
is used for semantic feature identification. These features
are concatenated and given as input to a SVM which
provides a probability distribution over all the predicate
classes. (see Fig.~\ref{fig3} and Fig.~\ref{fig4}).

\subsubsection{Object Detection} The first step of the solution approach is to detect the
objects present in an input image. Recent research works
have used various deep learning approaches and models.
These are developed in order to achieve high efficiency and
high accuracy for object detection. Approaches used in
literature include YOLO ~\cite{bib5}, R-CNN ~\cite{bib16}, Fast-RCNN ~\cite{bib17},
Faster-RCNN ~\cite{bib18}, Mask-RCNN ~\cite{bib20}, and Single-Shot
MultiBox Detector (SSD)~\cite{bib21}.

\begin{center}
{\includegraphics[width=7.5cm,height=10cm]{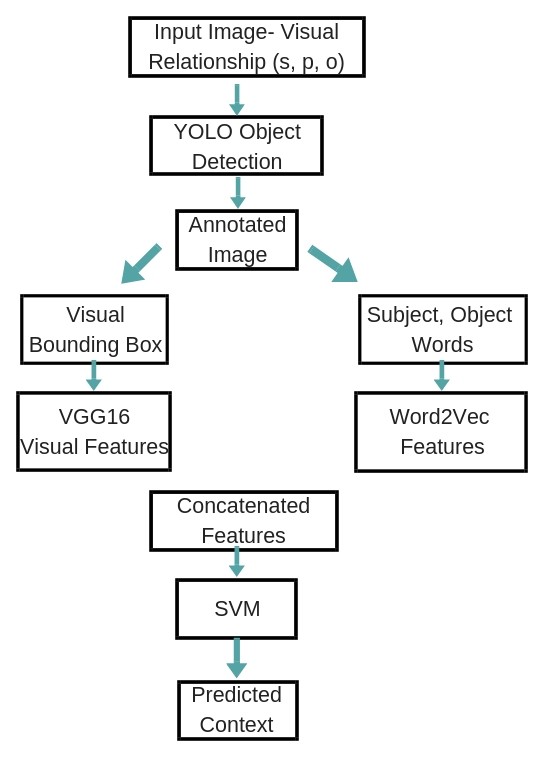}} \\
\textbf{Fig.3.}\text{ Flow Diagram of Solution Approach}
\label{fig3}
\end{center}

\noindent Here, YOLO (You Only Look Once) has been used for
object detection. It has an advantage that instead of using
local filters, it looks at an image globally and delivers
results. YOLO is very fast since it treats frame detection as
a regression problem. The model consists of CNN similar to
GoogleNet and instead of using residual block 1*1
convolutional layers are used. With 24 convolutional layers
and pre-trained on ImageNet dataset, the model is trained
for 135 epochs on PASCAL VOC dataset with a dropout of
0.5. Due to 1*1, the size of the prediction is the same as the
feature space. The dimension of the feature space is in the
format: 4 box coordinates, 1 objectness score, k class score
for every box on the image. Object score represents the
probability that an object is contained inside a bounding
box. Class confidences represent the probabilities of the
detected object belonging to a particular class. YOLO uses
softmax for class scores.

\subsection{Semantic and Visual Embedding}
Once the objects are detected, pairs for every object are
created giving us nC2 number of visual relationships. For
the visual features, the bounding box of subject and object
are taken. The predicate is the intersection over union (IoU)
of subject and object bounding boxes. All the three
bounding boxes are concatenated and given to a VGG16
network with predicate word as the ground truth label.
VGG16 is used for the classification task for the images.
It’s last layer provides good visual representations for the
objects in an image. Hence, it is extracted to get visual
relationship features for the concatenated bounding boxes.
Further, for the semantic embedding, Word2Vec is used
over the subject and object word. It is a powerful two layer
neural network that can convert text into a vector
representation. Word2Vec converts the subject and object
word into a 300 sized feature representation which is
concatenated and given to a neural network. The output
layer before the application of activation function is
extracted to get the semantic embedding for the visual
relationship. The generated semantic embedding are stored
in a dictionary format. The index is the object id and value
is the embedding. We store the object, predicate and their
embedding (found by word2vec) in the following formats
shown in Table~\ref{tab1}.

\begin{center}
\includegraphics[width=12cm,height=6cm]{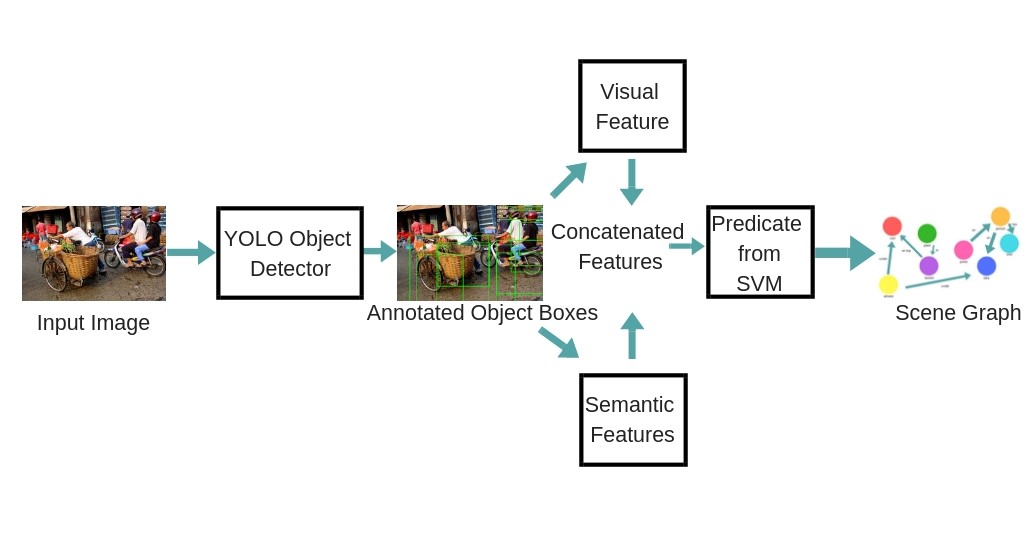}\\
\textbf{Fig.4.} \text{Phase Division for object detection and context interpretation}
\label{fig4}
\end{center}

% \begin{table}
% \caption{File details of object, predicate, and embedding}\label{tab1}
% \begin{tabular}{|l|l|l|}
% \hline
% File Name &  Function & Dictionary Format\\
% \hline
% Objects\_dict.pkl &  {Hashing of object} & Index-Object Name\\
% \hline
% Predicate\_dict.pkl &  {Hashing of predicates} & Index-Predicate Name\\
% \hline
% {Objects\_embedding.pkl} & {Objects Word2Vec embedding} & \makecell{}{object name-word2vec \\ & & embedding}\\
% \hline
% {Predicate\_embedding.pkl} & \makecell{Predicates Word2Vec embedding}& \makecell{predicate name-word2vec  \\ & & embedding}\\
% \hline
% \end{tabular}
% \end{table}

\begin{table}
\caption{File details of object, predicate, and embedding}\label{tab1}
\begin{tabular}{|l|l|l|}
\hline
File Name &  Function & Dictionary Format\\
\hline
Objects\_dict.pkl &  {Hashing of object} & Index-Object Name\\
\hline
Predicate\_dict.pkl &  {Hashing of predicates} & Index-Predicate Name\\
\hline
{Objects\_embedding.pkl} & {Objects Word2Vec embedding} &
\multirow{2}{*}{\begin{tabular}[c]{@{}l@{}}object name-word2vec\\embedding\end{tabular}} \\
& & \\
\hline
{Predicate\_embedding.pkl} & {Predicates Word2Vec embedding}& \multirow{2}{*}{\begin{tabular}[c]{@{}l@{}}predicate name-word2vec  \\ embedding\end{tabular}} \\
& & \\ 
\hline
\end{tabular}
\end{table}

\noindent Considering that the visual and semantic
embedding take the rich information about the image
which is not limited to only object detection, but also
to the semantic information present in an image, other
information which can also be taken is the spatial
feature representation which considers the location of
an object in an image with respect to the other objects.

\subsection{Predicate Detection}
The type of predicates in the dataset include the
spatial predicates like above, beside, on, etc.
Predicates can be of many types depicting the spatial
context and activity context like cooking, eating,
looking, etc. After the semantic and visual
embeddings are extracted, the embedding is
concatenated for a visual relationship in an image.
Some other methods can also be used when using
both the semantic and visual features which include,
multiplying both the feature, however, this requires
both the representation to be of the same size. The
dataset includes around 70 predicates. Since the
classes are quite distinct from each other, a decision
boundary between the classes would serve as a good
strategy to classify between the predicate classes and
SVM is a powerful discriminative model to achieve
this task. It is used as a classifier to give a class
distribution probability over all the 70 classes. The
class with the maximum probability is the predicted
class. For the scene graph, top 3 predicates are taken.
The predicate detected depicts the context shared
between the subject and object and thus delivers the
meaning of the image.

\subsection{Scene Graph Generation}
An image contains k number of visual relationships of
the format (subject, predicate, object). The predicate
was detected in the previous step. Now, the scene
graph is generated with nodes as objects/subjects and
edges as the predicate. Here, we use a directed scene
graph so that there is a differentiation between subject
and object. 

For example: for a statement, a person
eating food, the relationship format of (subject, predicate, object) would be (Person, eating,
Food). Here, the person is the subject, food is the
object, and eating is a predicate. If an undirected edge
is used, this statement loses the distinctive property of the person
being subject and food being object. The roles can be
reversed due to undirected edges leading to erroneous relationships. Therefore, the use
of directed edges is preferred. After the generation of
the scene graph, it can be traversed accordingly to
generate captions or summary of an image. The scene
graph can also be termed as a context graph.

\section{Findings}
An outcome for a sample VRD dataset image is shown in
Fig.~\ref{fig5}. The image after YOLO is shown in Fig.~\ref{fig6} which
shows the annotated image after YOLO object detection.
The objects detected in the shown image in the boundary
box are Person, wheel, cart, plant, bike, shirt, basket, and
pants. Mean Average Precision (MAP) is taken as a
performance measure to test the outcome. It considers the
average precision for recall of the detected objects and is a
popular metric for the object detectors. For the training set,
YOLO had an object detection accuracy of 55 MAP.

\begin{center}
\includegraphics[width=8cm,height=4.5cm]{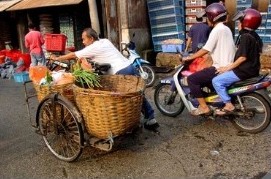}\\
\textbf{Fig.5.} \text{Input image from the VRD-Dataset} 
\label{fig5}
\end{center}

\begin{center}
\includegraphics[width=8cm,height=4.5cm]{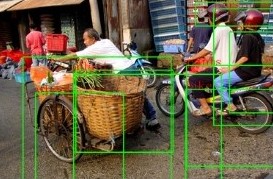}\\
\textbf{Fig.6.} \text{Annotated Image using YOLO Object Detector} \label{fig6}
\end{center}

\noindent Finally, a scene graph is generated based on these YOLO
detected visual features and semantic features. The scene
graph of Fig.~\ref{fig5} is depicted in Fig.~\ref{fig7}. Relationships
identified for which scene graph is formed are shown in
Table~\ref{tab2} showing the scene description using the subject predicate-object relationship.

The loss in the Neural network for a semantic feature
and CNN for the visual feature is shown in Fig.~\ref{fig8} and
Fig.~\ref{fig9} respectively. It is clearly observable in Fig.~\ref{fig8} that the training loss dropped with every epoch. The validation, however, increased after the 50th epoch more than the training. The point where the validation loss increases the training loss depicts the point where the model starts to
overfit. Hence, the weights of the network at the 50th epoch
were taken for further processing. One of the possible
reason of overfitting can be attributed to the dataset being
small. The model tries to fit to this small dataset and does
not learn the ability to generalize well.

\begin{center}
\includegraphics[width=7.5cm,height=5cm]{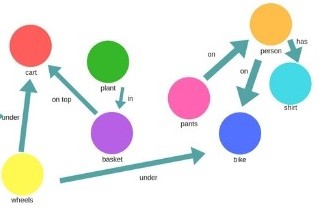}\\
\textbf{Fig.7.} \text{Scene Graph Result of the input image} \label{fig7}
\end{center}

\begin{table}
\centering
\caption{Details about scene, objects, and their relationships}\label{tab2}
\begin{tabular}{|l|l|l|}
\hline
Scene &  Relationships (s, p, o) \\
\hline
Wheel under cart &  (Wheel, under, cart) \\
\hline
Basket on top cart &  (Basket, on top, cart) \\
\hline
Plant in basket & (Plant, in, basket) \\
\hline
Wheel under bike & (Wheel, under, bike) \\
\hline
Pants on person & (Pants, on, person) \\
\hline
Person on bike & (Person, on, bike) \\
\hline
Person has shirt & (Person, has, shirt) \\
\hline
\end{tabular}
\end{table}

\begin{center}
\includegraphics[width=7.5cm,height=5.5cm]{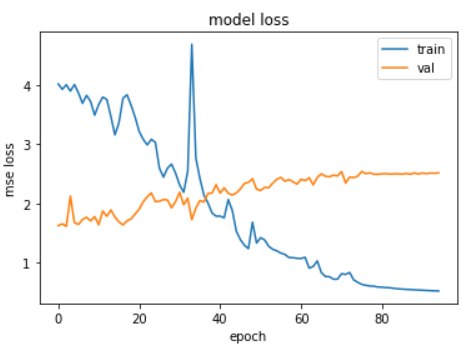}\\
\textbf{Fig.8.} \text{Training and Validation Loss for Visual Features}\label{fig8}
\end{center}

\begin{center}
\includegraphics[width=7.5cm,height=5.5cm]{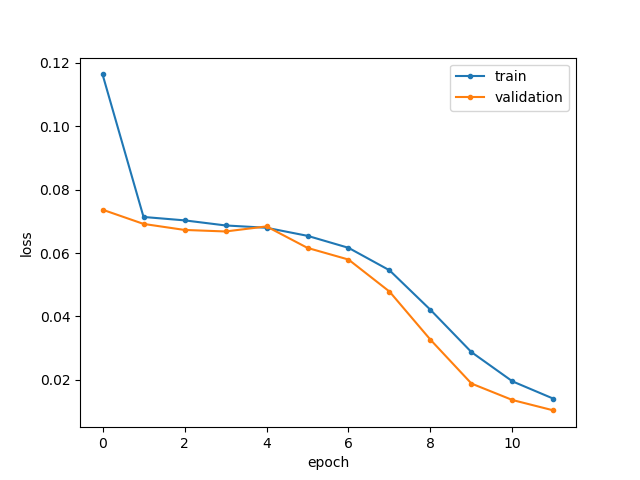}\\
\textbf{Fig.9.} \text{Training and Validation Loss for Semantic Features}\label{fig9}
\end{center}

\noindent The CNN and Neural Network were trained till they
reached an accuracy of 95\% and 99\% respectively. The
accuracy for predicate detection from SVM came out to be
60.57\%. The SVM was run for a total for 100 epochs. In our
previous approach of scene graph generation using
Word2vec solely, the accuracy reached till 40\% only.
However, once we incorporated the visual features also, the
accuracy increased to 60\%

\section{Conclusion and Future Scope}
Our work leverages the techniques of object detection by
finding out the context of the image in addition to the
detected object. We are detecting the context from the visual
and semantic features of an image. This is achieved by the
application of deep learning models YOLO for object
detection and Word2Vec for semantic feature representation
generation. A neural network is used for the semantic
feature of image and VGG16 for the visual feature
generation. Context can be used to find out the subtle
meaning of the image.
Future work includes extending the context to verbs like
cooking, eating, looking, etc since our work is covering only
the preposition predicates such as on, above, etc. Moreover,
in addition to the semantic and visual features, spatial
features can be incorporated which will be helpful in
determining the location of the objects. Lastly, better object
detection models like Faster-RCNN can be employed for
more accurate object detection in the first step because if an
object is not detected in the first step, it can't be used for
processing of visual relationships in the further stages.

\end{document}